\documentclass{article}

\pdfoutput=1
\usepackage[english]{babel}

\usepackage[letterpaper,top=2cm,bottom=2cm,left=3cm,right=3cm,marginparwidth=1.75cm]{geometry}

\usepackage{authblk}
\usepackage{amsmath}
\usepackage{amsthm}
\usepackage{graphicx}
\usepackage[colorlinks=false, allcolors=blue]{hyperref}
\usepackage[marginal]{footmisc}

\usepackage{epsfig}
\usepackage{graphicx}
\usepackage{mathrsfs}
\usepackage{color}
\usepackage{bm}
\usepackage{enumerate}
\usepackage{subfig}
\usepackage{booktabs}
\usepackage{threeparttable}
\usepackage{float}
\usepackage{epstopdf}
\usepackage{multirow} 
\usepackage{hyperref}

\newcommand{\hei}{\color{black}}
\def\bSig\mathbf{\Sigma}

\newcommand{\upcite}[1]{\textsuperscript{\textsuperscript{\cite{#1}}}}

    \author[1]{Yaohua Rong \thanks{Yaohua Rong and Sihai Dave Zhao are joint first authors.}}
    \author[2]{Sihai Dave Zhao}
    \author[1]{Xia Zheng}
    \author[3]{Yi Li \thanks{Corresponding author: Yi Li, Department of Biostatistics, University of Michigan, 1415 Washington Heights, Ann Arbor, MI 48109, USA. Email: yili@umich.edu}}
    \affil[1]{Faculty of Science, Beijing University of Technology, Beijing, Beijing, China}
    \affil[2]{Department of Statistics, University of Illinois at Urbana-Champaign, Champaign, Illinois, USA}
    \affil[3]{Department of Biostatistics, University of Michigan, Ann Arbor, Michigan, USA}

\begin{document}

\title{\bf Kernel Cox partially linear regression: building predictive models for cancer patients' survival}
\date{}

\maketitle

\begin{abstract}
Wide heterogeneity exists in cancer patients' survival, ranging from a few months to several decades. To accurately predict clinical outcomes, it is vital to build an accurate predictive model that relates patients' molecular profiles with patients' survival.  With complex relationships between survival and high-dimensional molecular predictors, it is  challenging to conduct non-parametric modeling and irrelevant predictors removing simultaneously. In this paper, we build a kernel Cox proportional hazards semi-parametric model and propose a novel regularized garrotized kernel machine (RegGKM) method to fit the model. We use the kernel machine method to describe the complex relationship between survival and predictors, while automatically removing irrelevant parametric and non-parametric predictors through a LASSO penalty. An efficient high-dimensional algorithm is developed for the proposed method. Comparison with other competing methods in simulation shows that the proposed method always has better predictive accuracy. We apply this method to analyze a multiple myeloma dataset and predict patients' death burden based on their gene expressions. Our results can help classify patients into groups with different death risks, facilitating treatment for better clinical outcomes. 

\end{abstract}

\section{Introduction}
There is wide variation in the survival of cancer patients, which can range from only a few months to several decades. Despite advancements in cancer treatment, there is still a lack of effective therapies for numerous types of cancer.In addition to designing effective cancer prevention programs, \upcite{barbeau2006results}  it is crucial to develop predictive models that robustly correlate patients' molecular profiles, \upcite{haddad2008hpv16} including genotypes and gene expressions, with their overall prognosis. This aligns  with precision medicine.  
 
With high-dimensional molecular predictors, regularization methods have been widely used for model selection and estimation. In survival settings, a family of penalized partial likelihood methods, such as the least absolute shrinkage and selection operator (LASSO)\upcite{Tibshirani1997the} and the smoothly clipped absolute deviation method (SCAD),\upcite{Fan2002Variable} has been proposed for Cox proportional hazards (PH) model. These methods select variables by shrinking some regression coefficients to exactly zero. For example,  Engler and Li \upcite{Engler2009Survival} adapt the elastic net approach to achieve variable selection for the Cox PH model, and Zhao and Li 
\upcite{zhao2012principled} propose a principled sure independence screening approach for 
selecting high dimensional predictors;  all of these works are based on a Cox PH model,  implicitly stipulating a linear relationship between predictors and log-hazard. As  the linear assumptions may not hold in some applications,  various nonparametric or partially linear models of predictors' effects have been proposed under the Cox PH framework.
Hastie and Tibshirani \upcite{Hastie1990Generalized} consider several step-wise nonlinear selection procedures; Du et al \upcite{Du2010PENALIZED} propose a joint estimation and selection method for the Cox partially linear model, in which the regression parameters in the parametric component are subjected to a SCAD penalty, while variable selection in the nonparametric component is performed in a nested fashion; Lian et al \upcite{Lian2013Shrinkage}  perform variable selection and estimation under an additive model.

Kernel machine has recently emerged as a powerful nonparametric method, by making only qualitative assumptions on functions, instead of a predetermined function or a nonparametric additive structure.  It is widely used for quantifying complex nonlinear and interactive effects, especially under high-dimensional settings. In ordinary regression, common kernel machine  approaches include Least Squares Kernel Machine,\upcite{liu2007semiparametric,liu2008estimation} Garrote Kernel Machine Testing,\upcite{maity2011powerful} as well as garrotized kernel-based variable selection  procedures,\upcite{Allen2013Automatic, He2016Prioritizing,Rong2018More,Zheng2021A} which have been developed for dimension reduction of the feature space.  With censored  outcomes, Leng and Zhang \upcite{Leng2006Model} extend the Component Selection and Smoothing Operator (COSSO) method \upcite{Lin2006Component} to the Cox PH model. The COSSO method is designed in the framework of smoothing splines ANOVA model and requires the use of univariate kernels, which may not sufficiently accommodate complex relationships.  Li and Luan \upcite{Li2003Kernel} estimate a fully nonparametric kernel Cox regression model by relating gene expression profiles with  survival response.  Sinnott and Cai \upcite{Sinnott2018Pathway}  propose a multiple kernel learning estimation method for censored survival outcomes. However, few of these works are available for variable selection,  partly due to the challenge of   nonparametric modeling with censored outcomes, especially under  high-dimensional scenarios.

To fill the gap, we propose a kernel Cox partially linear proportional hazards model and a novel  garrotized kernel machine method.  Our model flexibly depicts the possible complex relationship between  the response and  predictors while allowing  different non-parametric predictors to have different marginal effects.    We  design an efficient algorithm for implementing   the proposed method.  Our work is non-trivial and addresses several computational issues in practice. First, it is difficult to directly maximize the Cox PH objective function corresponding to the regularized garrotized kernel machine regression model. We tackle this by employing a quadratic approximation approach. Second, it is challenging to conduct efficient large-scale optimization in high-dimensional settings. We develop a spectral projected gradient method to handle this issue. Our method has been implemented in \textsf{R} (\textsf{R} Development Core Team 2020) and the code is available at \href{https://github.com/rongyaohua/RegGKM}{https://github.com/rongyaohua/RegGKM}.

The paper is structured as follows. Section 2  formulates the Cox PH kernel machine model,  introduces the  garrotized kernel-based estimation and variable selection method, and  proposes  an efficient computation algorithm for fitting the model.  Section 3 performs simulations to examine the finite sample performance of the proposed method, and Section 4 applies the proposed method to  a multiple myeloma dataset. Finally, Section 5 concludes the paper with  conclusions and the future work.

\section{METHODS}
\label{s:method}

\subsection{Regularized garrotized kernel machine for Cox partially linear model }
Denote by $D_i$ and $C_i$  the time from onset of one disease to death and the potential censoring time, respectively, for patient $i$, $i=1,\ldots,n.$ Let $ T_i = \min \{ D_i,C_i \}$ be the observed survival time, and $\tau_i = I(D_i \leq C_i)$ be the death indicator. For the $i$-th patient, let $\bm X_i = (X_{i1},\ldots,X_{iP})^T$  be a $P\times 1$ vector of clinical covariates such as age, and $\bm Z_i = (Z_{i1},\ldots,Z_{iQ})^T$ be a $Q\times 1$ vector of gene expressions. We assume independent censoring so that $D_i$ and $C_i$ are independent conditional on $\bm X_i$ and $\bm Z_i$. We adopt the following Cox partially linear proportional hazards model, where the conditional hazard function for the $i$-th patient with covariates $\bm X_i, \bm Z_i$ is 
\begin{equation}
	g(t|\bm X_i, \bm Z_i) = \underset{dt \rightarrow 0}{\lim} \frac{1}{dt} Pr(t \leq D_i < t+dt| D_i \geq t, \bm X_i, \bm Z_i)= g_0(t) \exp(\bm X_i^T \bm \beta + h(\bm Z_i)),
	\label{kercox}
\end{equation}
where $g_0(t)$ is the baseline hazard function,  $\bm \beta = (\beta_1,\ldots,\beta_P)^T$ quantifies the linear effects of nongenomic covariates on the (log) hazard, and $h(\cdot):\mathcal{R}^Q \rightarrow \mathcal{R}$ is an unknown function describing  possibly complicated relationships between genes and log-hazard.

We propose a kernel machine representation for $h(\cdot)$ by specifying that it lies in $\mathcal{H}_K$,  a reproducing kernel Hilbert space  ($\mathcal{RKHS}$) corresponding to the positive definite  kernel function $K(\cdot,\cdot): \mathcal{R}^Q \times \mathcal{R}^Q \rightarrow \mathcal{R} $. The kernel function $K(\bm Z_i,\bm Z_j) $ can be viewed as a measure  quantifying the similarity between gene profiles $\bm Z_i$ and $\bm Z_j$ for two subjects.    
The mathematical properties of $\mathcal{H}_K$ imply that any unknown function $h(\bm Z)$ in $\mathcal{H}_K$ can be represented as a weighted linear combination of the given kernel function $K(\cdot,\cdot)$ evaluated at each sample point: $h(\bm Z)=\sum^n_{i=1}K(\bm Z_i,\bm Z;\rho) \alpha_i$,  where the weights $\bm\alpha = (\alpha_1,\ldots,\alpha_n)$ is an $n\times 1$ \hei coefficient vector.\upcite{cristianini2000introduction,schaid2010genomic}

The choice of kernel function determines the particular function space where  the unknown smooth function $h({\cdot})$ is assumed to lie. Two popular kernel functions are the Gaussian  and  $d$-th polynomial kernels.  The Gaussian Kernel is defined as $K(\bm Z_1,\bm Z_2)=\exp\{-\sum_{q=1}^Q(Z_{1q}-Z_{2q})^2/\rho\}$, where $\rho$ is a tuning parameter. Its corresponding  $\mathcal{RKHS}$ contains many nonlinear functions \upcite{buhmann2003radial} and can approximate any continuous function.\upcite{Steinwart2002On}
The $d$-th polynomial kernel is $K(\bm Z_1,\bm Z_2)= (\bm Z_1^T \bm Z_2 +\rho )^d$, where both $\rho$ and $d$ are tuning parameters. The $d$-th polynomial kernel generates the function space spanned by all possible $d$-order monomials of the components of $\bm Z$. 
   We use  a set of functions $\phi(\bm Z)=\{Z_{k},Z_{k}Z_{k'}\}$  $(k,k^{'}=1,\ldots,Q)$  to represent the orthogonal basis functions that form an implicit orthogonal basis for the $\mathcal{RKHS}$ defined by the second-order polynomial kernel. In this model, the second-order polynomial terms are included, incorporating cross-product terms. \hei
Both Gaussian and the $d$-th $(d>1)$ polynomial kernels allow for gene-gene interactions.\upcite{liu2007semiparametric} There are other choices of kernel function, such as the neural network, smoothing spline \upcite{scholkopf2002learning} and identity-by-state kernels.\upcite{wu2011rare}
By choosing appropriate kernel functions, the complexity and form of the gene-risk function can be controlled. We use the Gaussian kernel to illustrate our new method, but in principle our approach is applicable to all of these kernel functions.

The unweighted kernel function mentioned above implicitly assumes that each nonparametric predictor has an equal effect on hazard. But in reality,  predictors in the nonparametric part may present different effects. To reflect this, we consider a garrotized kernel proposed by Rong et al,\upcite{Rong2018More} i.e.,
 given a base kernel $K(\cdot,\cdot)$, the garrotized kernel $K^{(g)}(\cdot,\cdot)$ is
\begin{equation}
	\begin{aligned}
		&K^{(g)}(\bm Z_i,\bm Z_j; \bm \delta)=K(\bm Z_i^*,\bm Z_j^*),\\
		& \bm Z_u^*=(\delta_1^{1/2}Z_{u1},\ldots,\delta_Q^{1/2}Z_{uQ})^T,u=i,j,\\
		&\delta_q\geq0,q=1,\ldots,Q.
	\end{aligned}
	\label{garker}
\end{equation}


For instance, the garrotized Gaussian kernel takes the form of $K^{(g)}( \bm Z_i,
\bm Z_j;\bm \delta)=\exp\{$ $-\sum_{q=1}^Q\delta_q(Z_{iq}-Z_{jq})^2\}$. The unknown parameter vector $\bm \delta$ needs to be estimated.  Each $\delta_q$ measures the effect of gene $Z_q$ on survival. For instance, $\delta_q=0$ means that the corresponding gene $Z_q$ is an irrelevant predictor, and, hence,  the garrotized kernel may offer a flexible means to identify the irrelevant predictive genes in $\bm Z$. 
Depending on the base kernel chosen $K(\cdot,\cdot)$, the function $h(\cdot)$ can still be  complicated and may not necessarily have an additive or ANOVA structure, for example allowing for complex high-order gene-gene interactions.

With predictors $\bm X$ and $\bm Z$  standardized to have  zero mean and unit variance,  
we propose to  maximize a regularized log-partial likelihood:
\begin{equation}
	\label{optpro}
	\begin{aligned}
		\underset{\bm \alpha,\bm \beta,\bm \delta}{\arg\,\max}
		\,
		\,  & \frac{1}{n} \sum_{i=1}^n \tau_i \left[\bm X_i^T \bm \beta+h(\bm Z_i)-\log\left\{\sum_{l\in R_i} \exp (\bm X_l^T\bm \beta+h(\bm Z_l))  \right\}  \right]        \\
		&
		\,- \lambda_1\sum_{p=1}^P\vert\beta_p\vert - \lambda_2\sum_{q=1}^Q\delta_q -
		\frac{1}{2}\lambda_3 \Vert h\Vert_{\mathcal{H}_{K^{(g)}}}^2,
	\end{aligned}
\end{equation}
where $R_i = \{l:T_l \geq T_i \}$ indexes patients at risk at time $T_i$,   
$\bm{\delta}: \delta_q\geq0,$ $q = 1,\ldots,Q,$ \hei
$\lambda_1$ and $\lambda_2$ are nonnegative regularization parameters, $\lambda_3 > 0$ is a tuning parameter  pursuing the trade-off between complexity of model and goodness of fit, and  $\Vert h\Vert_{\mathcal{H}_{K^{(g)}}}^2$ is
the functional norm defined in $\mathcal{H}_{K^{(g)}}$ generated by the garrotized kernel. Imposing a LASSO penalty on the parameters $\bm\beta$ and $\bm\delta$ may be suitable to accommodate sparsity, or only a small number of clinical and genetic predictors are actually associated with survival, while the penalty on $h(\cdot)$ is standard in kernel machine regression models.\upcite{Allen2013Automatic,Rong2018More,scholkopf2002learning}  

The proposed regularization (\ref{optpro}) enables estimation and variable selection of model \eqref{kercox} based on the garrotized kernel (\ref{garker}). The solution to this problem is given by Kimeldorf and Wahba\upcite{kimeldorf1970spline} via the representer theorem. That is, the optimal $h(\cdot)$ has the form of:
\begin{equation*}
	\begin{aligned}
		h(\bm Z)=\sum_{j=1}^n\alpha_j K^{(g)}(\bm Z, \bm Z_j; \bm\delta),
	\end{aligned}
\end{equation*}
where $K^{(g)}(\cdot,\cdot)$ is the garrotized kernel and $\alpha=(\alpha_1,\ldots,\alpha_n)^T$ is an unknown coefficient vector.
Consequently, maximization of~\eqref{optpro} is equivalent to maximizing
\begin{equation}
	\label{max}
	\begin{aligned}
		\, f(\bm \alpha,\bm \beta,\bm \delta) =
		\,  & \frac{1}{n} \sum_{i=1}^n \tau_i \left[\bm X_i^T \bm \beta+ \bm k_i(\bm \delta) \bm\alpha - \log\left\{\sum_{l\in R_i} \exp (\bm X_l^T\bm \beta + \bm k_l(\bm \delta) \bm\alpha )  \right\}  \right]        \\
		&
		- \lambda_1\sum_{p=1}^P\vert\beta_p\vert - \lambda_2\sum_{q=1}^Q\delta_q -
		\frac{1}{2}\lambda_3\bm\alpha^T \bm K(\bm\delta)\bm\alpha,
	\end{aligned}
\end{equation}
where $\bm k_i(\bm \delta) = (K^{(g)}(\bm Z_i,\bm Z_1;\bm\delta),\ldots,K^{(g)}(\bm Z_i,\bm Z_n;\bm\delta))$, and $\bm K(\bm \delta)$ is an $n\times n$  symmetric Gram matrix, with the $ij$-th element of $\bm K_{ij}(\bm \delta)=K^{(g)}(\bm Z_i,\bm Z_j;\bm \delta)$, $i,j=1,\ldots, n$.
Equivalently, the  objective function \eqref{max} can be written as
\begin{equation}
	\label{maxvec}
	\begin{aligned}
		\, f(\bm \alpha,\bm \beta,\bm \delta) =
		\,  & \frac{1}{n} \bm\tau^T \left[\bm X \bm\beta+ \bm K(\bm\delta)\bm\alpha - \bm A(\bm\alpha,\bm\beta,\bm\delta)  \right]        \\
		&
		- \lambda_1 \Vert\bm\beta\Vert_1 - \lambda_2\Vert\bm\delta\Vert_1 -
		\frac{1}{2}\lambda_3\bm\alpha^T \bm K(\bm\delta)\bm\alpha,
	\end{aligned}
\end{equation}
where $\bm\tau=(\tau_1,\ldots,\tau_n)^T$ is a $n\times 1$  death indicator vector, $\bm X=(\bm X_1,\ldots,\bm  X_n)^T$ is the $n\times P$ clinical covariate matrix, $\bm A(\bm\alpha,\bm\beta,\bm\delta)=\{\log[\underset{l\in R_1}{\sum} \exp (\bm X_l^T\bm \beta + \bm k_l(\bm \delta) \bm\alpha) ],\ldots$, $\log[\underset{l\in R_n}{\sum} \exp (\bm X_l^T\bm \beta + \bm k_l(\bm \delta) \bm\alpha) ]  \}^T$ is an $n\times 1$  vector, $\Vert\cdot\Vert_1$ is the $L_1$ norm. 

We refer to the solution of equation~\eqref{maxvec} as the Regularized Garrotized Kernel Machine estimate (RegGKM) for the Cox partially linear proportional hazards model. Our proposed method, as outlined in the following section, does not impose any restrictions on the number of genomic covariates that can be used in the model.

\subsection{Algorithm}
\label{algo}
We propose a group coordinate descent algorithm to maximize  \eqref{maxvec}  to estimate the parameters of $\bm\alpha,\bm\beta,\bm\delta$, which can be computed along a regularization path as follows.  

\begin{enumerate}
	\item[Step 1.]    Set initial estimates $\bm\alpha^0$,$\bm\beta^0$,$\bm\delta^0$. 
 \footnotetext{   
 To investigate the impact of initial values on the algorithm, we have conducted simulation experiments. The results demonstrate that the proposed algorithm is not sensitive to initial values. We select $(1/Q, \ldots, 1/Q)$ and $(1/n, \ldots, 1/n)$ as the initial values for $\bm\delta$ and $\bm\alpha$, respectively. The initial value of $\bm\beta$ is the LASSO estimate obtained by fitting a penalized linear COX model.}\hei
	\item[Step 2.] Update $\bm\alpha,\bm\beta,\bm\delta$ cyclically    for a given regularization parameter $\bm\lambda = (\lambda_1,\lambda_2,\lambda_3)$. \hei Specifically, 
	
	\begin{itemize}
		
		\item With the values of $\bm\alpha,\bm\delta$ fixed at $\tilde{\bm\alpha},\tilde{\bm\delta}$, the objective function \eqref{maxvec} can be written as
		\begin{equation*}
			\begin{aligned}
				\, f(\tilde{\bm\alpha},\bm \beta,\tilde{\bm\delta}) =
				\,  & \frac{1}{n} \bm\tau^T \left[\bm X \bm\beta+ \bm K(\tilde{\bm\delta})\tilde{\bm\alpha} - \bm A(\tilde{\bm\alpha},\bm\beta,\tilde{\bm\delta})  \right]        \\
				& - \lambda_1 \Vert\bm\beta\Vert_1 - \lambda_2\Vert\tilde{\bm\delta}\Vert_1 -
				\frac{1}{2}\lambda_3\tilde{\bm\alpha}^T \bm K(\tilde{\bm\delta})\tilde{\bm\alpha},
			\end{aligned}
		\end{equation*}
		
		 As it may be difficult to directly maximize this nonconvex objective function,  we propose to use a  quadratic function to  approximate it locally and  adopt a penalized reweighted least squares approach to update $\bm\beta$.\upcite{Tibshirani1997the}  Denote the log-partial likelihood as $\ell_n(\bm\alpha,\bm\beta,\bm\delta) = \frac{1}{n} \bm\tau^T [\bm X \bm\beta+ \bm K(\bm\delta)\bm\alpha - \bm A(\bm\alpha,\bm\beta,\bm\delta) ]$ and define $\bm\eta = \bm X \bm\beta+ \bm K(\bm\delta)\bm\alpha$. Let $\ell_n^{'}(\bm\eta)$ and $\ell_n^{''}(\bm\eta)$ be the gradient and Hessian of the log-partial likelihood with respect to $\bm\eta$, respectively; see their explicit expressions  in the Appendix. 
		Given the current estimate $\tilde{\bm\eta} = \bm X \tilde{\bm\beta}+ \bm K(\tilde{\bm\delta})\tilde{\bm\alpha}$, maximizing log-partial likelihood $\ell_n(\bm\eta)$ is equivalent to maximizing a two-term Taylor expansion of the log-partial likelihood
		\begin{equation*}
			\frac{1}{2n}[\bm Y(\tilde{\bm\eta}) - \bm X \bm\beta - \bm K(\tilde{\bm\delta})\tilde{\bm\alpha}]^T \ell_n^{''}(\tilde{\bm\eta}) [\bm Y(\tilde{\bm\eta}) - \bm X \bm\beta - \bm K(\tilde{\bm\delta})\tilde{\bm\alpha}],
		\end{equation*}
		
		where $\bm Y(\tilde{\bm\eta}) = \tilde{\bm\eta} - \ell_n^{''}(\tilde{\bm\eta})^{-1}\ell_n^{'}(\tilde{\bm\eta})$. Similar to Tibshirani,\upcite{Tibshirani1997the}  to avoid computational difficulty, we approximate the Hessian matrix with a diagonal one, e.g., setting off-diagonal elements to zero, and minimize the penalized reweighted least squares w.r.t. $\bm\beta$
		\begin{equation*}
			\frac{1}{2n}[\bm Y(\tilde{\bm\eta}) - \bm X \bm\beta - \bm K(\tilde{\bm\delta})\tilde{\bm\alpha}]^T [\bm W(\tilde{\bm\eta})] [\bm Y(\tilde{\bm\eta}) - \bm X \bm\beta - \bm K(\tilde{\bm\delta})\tilde{\bm\alpha}] + \lambda_1\Vert\bm\beta\Vert_1,
		\end{equation*}
		where $\bm W(\tilde{\bm\eta})$ is a diagonal matrix that has the same diagonal elements as $- \ell_n^{''}(\tilde{\bm\eta})$. Hence, treating $\bm Y(\tilde{\bm\eta}) - \bm K(\tilde{\bm\delta})\tilde{\bm\alpha}$ as the working response, $\bm W(\tilde{\bm\eta})$ as weight, standard procedures for solving LASSO regression estimates could be used to estimate the $\bm\beta$.\upcite{friedman2007pathwise,friedman2010regularization}
		
		\item Fixing $\bm\beta,\bm\delta$ at values $\tilde{\bm\beta},\tilde{\bm\delta}$, we update $\bm\alpha$. The maximization problem \eqref{maxvec} can be expressed as
		\begin{equation*}
			\label{updaph}
			\begin{aligned}
				\, f(\bm\alpha,\tilde{\bm\beta},\tilde{\bm\delta}) =
				\,  & \frac{1}{n} \bm\tau^T \left[\bm X \tilde{\bm\beta} + \bm K(\tilde{\bm\delta})\bm\alpha - \bm A(\bm\alpha,\tilde{\bm\beta},\tilde{\bm\delta})  \right]        \\
				& - \lambda_1 \Vert\tilde{\bm\beta}\Vert_1 - \lambda_2\Vert\tilde{\bm\delta}\Vert_1 -
				\frac{1}{2}\lambda_3\bm\alpha^T \bm K(\tilde{\bm\delta})\bm\alpha.
			\end{aligned}
		\end{equation*}
		Similar to the updating procedure of $\bm\beta$, maximizing the previous function is equivalent to minimizing the following function w.r.t $\bm\alpha$,
		\begin{equation*}
			\frac{1}{2n}[\bm Y(\tilde{\bm\eta}) - \bm X\tilde{\bm\beta} - \bm K(\tilde{\bm\delta})\bm\alpha]^T \bm W(\tilde{\bm\eta}) [\bm Y(\tilde{\bm\eta}) - \bm X\tilde{\bm\beta} - \bm K(\tilde{\bm\delta})\bm\alpha] +
			\frac{1}{2}\lambda_3\bm\alpha^T \bm K(\tilde{\bm\delta})\bm\alpha,
		\end{equation*}
		which is a quadratic function of $\bm\alpha$. It follows that the update for $\bm\alpha$ is the solution to a linear system of equations
		\[
		\left[\frac{1}{n} \bm K({\tilde{\bm\delta}}) \bm W(\tilde{\bm\eta}) \bm K({\tilde{\bm\delta}}) + \lambda_3\bm K({\tilde{\bm\delta}})\right] \bm\alpha=\frac{1}{n} \bm K({\tilde{\bm\delta}}) \bm W(\tilde{\bm\eta}) (\bm Y(\tilde{\bm\eta})- \bm X{\tilde{\bm\beta}}).
		\]
		    If the left-hand side of the above equation becomes a singular matrix, it is necessary to add a diagonal matrix with small entries, such as $0.00001$, to stabilize the estimate. \hei

		\item For $\bm \delta$, given the estimates of $\alpha$, $\beta$, \eqref{maxvec} can be written as 
		\begin{equation*}
			\label{upddel}
			\begin{aligned}
				\, f(\tilde{\bm\alpha},\tilde{\bm\beta},\bm\delta) =
				\,  & \frac{1}{n} \bm\tau^T \left[\bm X \tilde{\bm\beta} + \bm K(\bm\delta)\tilde{\bm\alpha} - \bm A(\tilde{\bm\alpha},\tilde{\bm\beta},\bm\delta)  \right]        \\
				& - \lambda_1 \Vert\tilde{\bm\beta}\Vert_1 - \lambda_2 \Vert\bm\delta\Vert_1-
				\frac{1}{2}\lambda_3\tilde{\bm\alpha}^T \bm K(\bm\delta)\tilde{\bm\alpha}.
			\end{aligned}
		\end{equation*}

        With the updated values of $\bm\beta,\bm\alpha$,   say,  $\tilde{\bm\beta},\tilde{\bm\alpha}$, updating $\bm\delta$ can be done by  solving a nonlinear optimization problem under the constraint of $\bm\delta>\bm 0$.
        To address the difficulty in high-dimensional optimization, we propose using the spectral projected gradient (SPG) method, a non-monotone projected gradient algorithm. This method is well-suited for solving large-scale convex-constrained optimization problems in a more efficient manner when gradients are readily computable,
        as is the case for the gradient of the objective function with respect to $\bm\delta$, as detailed in the Appendix.   Additionally, its implementation does not require the Hessian matrix, making it ideal for high-dimensional problems with thousands of predictors.
        We implement the SPG method  by using the function \texttt{SPG} in the \textsf{R} package of \textbf{BB}.\upcite{2009BB} 
	\end{itemize}
	
	\item[Step 3.]  Repeat Step 2 until convergence    or the number of iteration reaches a pre-specified maximum number of iterations . \hei
\end{enumerate}

\subsection{Selection of tuning parameters}

Once the solution path has been determined, it is necessary to select the optimal set of regularization/tuning parameters, denoted as $\bm\lambda = (\lambda_1, \lambda_2, \lambda_3)$. The selection of $\bm\lambda$ is an important aspect within the kernel machine framework. Our proposed RegGKM procedure  is relatively new  and there is no established criterion available for selecting the tuning parameters. We suggest using cross-validated partial log-likelihood \upcite{Van2006Cross} to determine the optimal tuning parameters.

In general, tuning parameters can be selected using $K$-fold cross validation. This process involves dividing the data into $K$ pieces, training the model using $K-1$ pieces, and validating on the $k$-th piece, as described by Friedman et al \upcite{friedman2010regularization} and Van Houwelingen et al \upcite{Van2006Cross}. This procedure is repeated, validating on each of the $K$ pieces in turn, and finally, the $K$ deviances are summed. However, as noted by Friedman et al,\upcite{friedman2010regularization} cross-validation in the Cox model presents some subtle differences. For example, when using leave-one-out cross-validation, the predictive partial likelihood for the left-out observation is either identically 1 for all $\eta$ (if the left-out observation is an observed event) or undefined (if the left-out observation is right-censored).
To fix this ill-conditioned problem, we use the cross-validated partial log-likelihood (CVPL) scheme proposed by Van Houwelingen et al.  For the $k$-th fold, denote the leave-one-fold-out estimates as $\hat{\bm\beta}_{-k}(\bm\lambda)$, $\hat{\bm\delta}_{-k}(\bm\lambda)$, $\hat{\bm\alpha}_{-k}(\bm\lambda)$, which are computed by maximizing the leave-one-fold-out penalized partial log-likelihood $\ell_{-k}(\bm\alpha,\bm\beta,\bm\delta) + \lambda_1 \Vert\bm\beta\Vert_1 + \lambda_2\Vert\bm\delta\Vert_1 +
\frac{1}{2}\lambda_3\bm\alpha^T \bm K(\bm\delta)\bm\alpha$ based on non-left-out data for a given $\bm\lambda$.  \upcite{Van2006Cross}\upcite{Simon2011Regularization} Our goodness of fit for a given $\bm\lambda$ can be measured by 
\begin{equation}
	\text{CVPL}(\bm\lambda) = \sum_{k=1}^{K}\{ \ell(\hat{\bm\beta}_{-k}(\bm\lambda),\hat{\bm\delta}_{-k}(\bm\lambda),\hat{\bm\alpha}_{-k}(\bm\lambda)) - \ell_{-k}(\hat{\bm\beta}_{-k}(\bm\lambda),\hat{\bm\delta}_{-k}(\bm\lambda),\hat{\bm\alpha}_{-k}(\bm\lambda)) \}.
\end{equation}

{   For a given regularization path $\Lambda=(\bm\lambda_1, \bm\lambda_2, \ldots)$, we compute the estimates of $\bm\alpha(\bm\lambda)$, $\bm\beta(\bm\lambda)$, and $\bm\delta(\bm\lambda)$ for each $\bm\lambda$ in $\bm\Lambda$ and obtain a solution path. Finally, we maximize the proposed cross-validated partial log-likelihood (CVPL) with respect to $\bm\lambda$ to obtain the optimal tuning/regularization parameters, which correspond to the optimal estimates. Specifically, the regularization parameter $\bm\lambda$ in the second step of the algorithm is initially selected from a large range and gradually narrowed down based on the proposed CVPL until the optimal regularization parameter is accurately selected within a small range.} 

\section{Simulation}
\label{s:sim}
\subsection{Prediction  measures}
In simulations and real data analysis, we use CVPL, along with C-statistics and area under the time-dependent receiver operating characteristic (ROC) curves (AUC), to compare our proposed RegGKM method with other competing methods. The latter two metrics are defined as follows, and are calculated by  \textsf{R} packages ``{survC1}'', ``{survivalROC}'' and ``{risksetROC}''.

(i) \emph{C-statistic}.
For a specified follow-up period, say, $(0,\xi)$,  the  C-statistic  is defined as \upcite{uno2011on}
\begin{equation*}
	\begin{aligned}
		C_{\xi}=Pr(f(X_1, Z_1)>f(X_2, Z_2)|D_2>D_1, D_1<\xi),
	\end{aligned}
\end{equation*}
where  $f(X,Z)=X^T  \hat{\beta} + \hat{h}(Z)$ is the estimated risk score by the RegGKM method. When $D$ is right censored, Uno et al\upcite{uno2011on} account for censoring via  inverse probability weighting,  obtaining a  censoring-adjusted estimate of $C_{\xi}$:
\begin{equation*}
	\begin{aligned}
		\hat{C}_{\xi}=\frac{\sum_{i=1}^{n}\sum_{j=1}^{n}\tau_i\{\hat{G}(T_i)\}^{-2}I(T_i<T_j,T_i<\xi)I(X_i^T  \hat{\beta} + \hat{h}(Z_i)>X_j^T  \hat{\beta} + \hat{h}(Z_j))}{\sum_{i=1}^{n}\sum_{j=1}^{n}\tau_i\{\hat{G}(T_i)\}^{-2}I(T_i<T_j,T_i<\xi)},
	\end{aligned}
\end{equation*}
where $I(\cdot)$ is the indicator function and $\hat{G}(\cdot)$ is the Kaplan-Meier estimator for censoring distribution $G(t)=Pr(C>t)$.
As for the choice of $\xi$,  Uno et al.\upcite{uno2011on} recommend to choose
$\xi$ to balance estimation accuracy and numerical stability. Following Uno et al.\upcite{uno2011on}, we have chosen $\xi$ to be the $70$ percentile of observed survival time when computing C-statistics in the simulations and in the real data analysis.

(ii) \emph{AUC}. 
The ROC curve displays the true positive rate (sensitivity) versus the false positive rate (1-specificity) for all possible cutpoints, providing  all possible combinations of true-positive and false-positive rates. In a survival setting,  the AUC at $t$, i.e., AUC($t$), is equal to the probability that the marker value of a randomly chosen diseased subject is above the marker value of a randomly chosen  subject alive at that time, where the marker value is equal to $X^T \hat{\beta} + \hat{h}( Z)$ for the RegGKM method,  and can also be viewed as a concordance index. The overall AUC over  $(0,\xi)$ is the weighted integral of AUC($t$) over  $(0,\xi)$.
\upcite{heagerty2005survival}
 As in the calculation of the C-statistic,  the choice of $\xi$  entails a delicate balance between achieving accurate estimations and maintaining numerical stability. Following the guidance provided by Heagerty and Zheng\upcite{heagerty2005survival} and based on our own experience,  we have chosen $\xi$ to be the 90\% of the maximum observed survival time when computing the overall AUC in the simulations and in the real data analysis.

\hei
\subsection{Comparison with LASSO-COX}
\label{comlasso}

To investigate the finite sample performance, we firt compare our proposed RegGKM method with the LASSO-COX method.
The data are generated from the following Cox partially linear  proportional hazards model,
\begin{equation}
	g(t| X_i,  Z_i) = \exp( X_i^T  \beta + h( Z_i)),
	\label{simulation}
\end{equation}
where the non-genomic covariates $X_{ip}$ and genomic covariates $Z_{iq}$ are independently generated from $U(-0.01, 0.01)$ and $U(0, 3)$, respectively. To imitate the complex relationships between gene expression levels and the hazard, the nonparametric function $h(\cdot)$ is set to have a complex form with nonlinear functions of the $Z$'s and interactions among the $Z$'s.
In practice, without prior knowledge, the true relevant predictors are not known in advance. The predictor set used in modeling is always larger than the true relevant predictor set. To mimic such a scenario, we consider the following 7 settings, where Settings 1-2,  3-5 and  6-7 represent the multivariate, high-dimensional and ultrahigh-dimensional cases, respectively.

\begin{enumerate}
\item[1:] $P=1$, $Q=5$, 
$\beta=1$, $h(Z)=0.6\cos(Z_1)Z_2+0.36Z_1^2 - 0.3\exp(Z_1)Z_2 -0.36\sin(Z_2)\cos(Z_3) +0.6\exp(Z_3)\sin(Z_4) -0.48Z_2\sin(Z_4)-0.12\cos(Z_3)Z_4^2 -0.12\exp(Z_4)\cos(Z_5)-0.48\sin(Z_4)Z_5^2.$ We fit Model~\eqref{simulation} without any additional irrelevant predictors.

\item[2:] $P=2$, $Q=15$, $ \beta=(1,0)^T$, $h(\cdot)$ is the same as in Setting 1. We fit Model~\eqref{simulation} with 1 additional irrelevant $X$ predictor and 10 additional irrelevant $Z$ predictors.

\item[3:] $P=200$, $Q=15$, $ \beta=(\beta_1^T, \beta_0^T)^T$, where $\beta_1= (1,1,1,1,1)^T$ and $\beta_0$ is the zero vector, $h(Z)=0.72\cos(Z_1)Z_2 - 0.24\exp(Z_1)Z_2 +0.72\exp(Z_2)\sin(Z_3)-0.12\cos(Z_1)Z_3^2 -0.12\exp(Z_2)\cos(Z_3)$. We fit Model~\eqref{simulation} with 195 additional irrelevant $X$ predictors and 12 additional irrelevant $Z$ predictors.

\item[4:] $P=15$, $Q=200$, $ \beta=(\beta_1^T, \beta_0^T)^T$, where $\beta_1= (1,1,1,1,1)^T$ and $\beta_0$ is the zero vector, $h(\cdot)$ is the same as in Setting 3. We fit Model~\eqref{simulation} with 10 additional irrelevant $X$ predictors and 197 additional irrelevant $Z$ predictors.

\item[5:] $P=200$, $Q=200$, $ \beta=(\beta_1^T, \beta_0^T)^T$, where $\beta_1= (1,1,1,1,1)^T$ and $\beta_0$ is the zero vector, $h(\cdot)$ is the same as in Setting 3. We fit Model~\eqref{simulation} with 195 additional irrelevant $X$ predictors and 197 additional irrelevant $Z$ predictors.

\item[6:] $P=1$, $Q=1000$, $ \beta=1$, $h(\cdot)$ is the same as in  Setting 3. We fit Model~\eqref{simulation} with 997 additional irrelevant $Z$ predictors.

\item[7:] $P=1000$, $Q=1000$, $ \beta=(\beta_1^T, \beta_0^T)^T$, where $\beta_1= (1,1,1,1,1)^T$ and $\beta_0$ is the zero vector, $h(\cdot)$ is the same as in Setting 3. We fit Model~\eqref{simulation} with 995 additional irrelevant $X$ predictors and 997 additional irrelevant $Z$ predictors.
\end{enumerate}
The censoring time is generated from an exponential distribution with mean $U\exp( X^T \beta + h( Z))$, where $U$ is  randomly generated from a uniform distribution. The different censor rates  depend on different uniform distributions from which $U$ is randomly generated. For each setting, only the garrotized Gaussian kernel is used to fit model (\ref{simulation}). We take the sample size $n = 100$  and conduct a total of    1,000 \hei experiments for each parameter configuration.

The CVPLs, C-statistics and AUCs of RegGKM and LASSO-COX for the seven settings under different censor rates are reported in Table~\ref{mulvar} (multivariate settings) and Table~\ref{higdim} (high-dimensional and ultrahigh-dimensional settings). The simulation results clearly suggest that our proposed RegGKM method  outperforms the LASSO-COX in CVPL, C-statistics and AUC  under all of the scenarios examined.  The RegGKM method has a better prediction performance compared to the LASSO-COX method,  possibly because the proposed RegGKM method is more  model-free, while the LASSO-COX method is sensitive to the misspecified model when the data are generated from a Cox partially linear model. This may suggest that, if there are complex nonlinear structures and gene-gene interactions, the RegGKM method could capture these effects better than the LASSO-COX method and hence gains more accuracy.

\subsection{Comparison with COSSO-COX}

To evaluate the performance of our proposed RegGKM method against the COSSO-COX method, we fit a Cox partially linear hazards model based on the RegGKM method, and a nonparametric model, an additive ANOVA model including only the main effects, using the COSSO-COX method. As the COSSO-COX method is not suitable for high-dimensional settings, we only considered Settings 1 and 2 in Section~\ref{comlasso} and conducted a total of 100 replications for each parameter configuration.

Table~\ref{prediction4} presents a comparison of the predictive performance of these two methods, considering two settings with different censor rates. The results clearly demonstrate that our proposed RegGKM method surpasses the COSSO-COX method in terms of average CVPLs, C-statistics, and AUCs. This is likely due to the fact that the RegGKM method allows for complex relationships, including gene-gene interactions, while the COSSO-COX method only considers main effects in its additive model.

\section{Analysis of the multiple myeloma dataset}

 We applied the RegGKM method to study multiple myeloma using data from the Gene Expression Omnibus (GSE24080). The dataset contained 20162 genes from 169 patients, including 125 patients who were observed to die during followup and 44 patients who were censored.
Our goal was to develop a more accurate predictive model for multiple myeloma patients' survival that takes into account both gene expression measurements and clinical information. To do this, we first conducted a preliminary screening of genes based on existing findings\upcite{broyl2010gene}  and retained 129 genes deemed most relevant to survival outcomes. Then, we applied the proposed RegGKM method, incorporating five linearly modeled clinical features (age, creatinine, lactate dehydrogenase, albumin and percent of plasma cells in a bone marrow biopsy) and 129 nonparametrically modeled genes. 
   In general, the selection of a kernel function can be guided by subject matter knowledge or the specific signals researchers wish to consider, such as linear or nonlinear effects. Since our focus was on gene expression data, we recommend using the garrotized Gaussian kernel, which allows for flexible capture of various linear and nonlinear effects in gene expression data.\upcite{Sinnott2018Pathway}  It is worth noting that different types of genomic data may have natural kernel choices. For instance, the identity-by-state kernel has been frequently used for genotype data. \hei  Additionally, we compared the results with those obtained using the COSSO-COX and LASSO-COX methods. 

To evaluate the predictive accuracy of the three methods, we randomly divided the observations into a training set (85 observations), a validation set (42 observations) for finding the optimal tuning parameters, and a testing set (the remaining observations) for prediction evaluation. The average cross-validated partial likelihood (CVPL), C-statistics, and area under the curve (AUC) for each method were calculated over 500 replications and reported in Table~\ref{realtable}. As shown in the table, our RegGKM method demonstrated superior predictive accuracy compared to the other two methods. The genes selected by the RegGKM method can provide important information about the biological processes underlying multiple myeloma. For example, FCGR2B (Fc fragment of IgG receptor IIb), FGFR3 (fibroblast growth factor receptor 3) and DUSP5 (dual-specificity phosphatase 5) were the top three genes selected most frequently by RegGKM over 500 replications. FCGR2B was found to be negatively correlated with the overall survival rate in patients with clear cell renal cell carcinoma,\upcite{li2021bioinformatics} FGFR3 is mutated or overexpressed in many bladder cancers,\upcite{williams2013oncogenic} and DUSP5 has been implicated in skin carcinogenesis. \upcite{rushworth2014dual}

The LASSO-COX approach also selected genes: CD163 (CD163 molecule), KIT (proto-oncogene, receptor tyrosine kinase) and DSG2 (desmoglein 2). These differ from those selected by the RegGKM method and may provide complementary information about the biology of multiple myeloma. However, the improved predictive performance of RegGKM compared to the LASSO-COX and COSSO-COX methods highlights the benefits of our proposed method, which accommodates complex interactions and relationships between genes and hazards. Our results suggest that a simple linear assumption is not flexible enough to capture the complex relationships between genes, hazard, and gene-gene interactions.

\section{Conclusion}
\label{s:discuss}
To improve the prediction accuracy of the Cox proportional hazard model when dealing with the complex relationships between response and predictors in high-dimensional settings, we constructed a kernel Cox regression model by introducing the regularized garrotized kernel machine method and designed an efficient algorithm for implementation. A vital advantage of the proposed RegGKM method is that it  not only captures the complex relationships between microarray gene expression profiles and hazard and allows for possible interactions among genes, but also removes the irrelevant predictors automatically and hence possesses more predictive accuracy compared to the competing methods. Several simulations and the analysis of the MM patients survival data reveal that the predictive superiority of our proposed RegGKM method to the existing methods.

It is worth mentioning that we considered only kernel Cox proportional hazard model using a garrotized Gaussian kernel. Of course, the proposed garrotized kernel machine procedure can also be extended to other survival models, such as accelerated failure model. Besides, the RegGKM method can also accommodate more kinds of data, such as longitudinal and missing data. We are also pursuing extensions to quantile regression and will present the results elsewhere.



\section*{ACKNOWLEDGEMENTS}
Rong’s work is supported by the National Natural Science Foundation of
China (No.11701021), National Statistical Science Research Project (No.2022LZ22), Science and Technology Program of Beijing Education Commission (No.KM202110005013),
and National Natural Science Foundation of China (No.11971001).
\vspace*{-8pt}

\subsection*{FINANCIAL DISCLOSURE}

None reported.

\subsection*{CONFLICTS OF INTEREST}

The authors declare no potential conflict of interests.

\label{lastpage}

\bibliographystyle{unsrt}

\bibliography{ref}%

\begin{thebibliography}{10}

\bibitem{barbeau2006results}
Elizabeth~M Barbeau, Yi~Li, Patricia Calderon, Cathy Hartman, Margaret Quinn,
  Pia Markkanen, Cora Roelofs, Lindsay Frazier, and Charles Levenstein.
\newblock Results of a union-based smoking cessation intervention for
  apprentice iron workers.
\newblock {\em Cancer Causes \& Control}, 17:53--61, 2006.

\bibitem{haddad2008hpv16}
Robert Haddad, Christopher Crum, Zigui Chen, Jeffrey Krane, Marshall Posner,
  Yi~Li, and Robert Burk.
\newblock Hpv16 transmission between a couple with hpv-related head and neck
  cancer.
\newblock {\em Oral Oncol.}, 44(8):812--815, 2008.

\bibitem{Tibshirani1997the}
Robert Tibshirani.
\newblock The lasso method for variable selection in the cox model.
\newblock {\em Stat Med.}, 16(4):385--395, 1997.

\bibitem{Fan2002Variable}
Jianqing Fan and Runze Li.
\newblock Variable selection for cox's proportional hazards model and frailty
  model.
\newblock {\em Ann Stat.}, 30(1):74--99, 2002.

\bibitem{Engler2009Survival}
D~Engler and Y.~Li.
\newblock Survival analysis with high-dimensional covariates: an application in
  microarray studies.
\newblock {\em Stat Appl Genet Mol Biol.}, 8(1):1--22, 2009.

\bibitem{zhao2012principled}
Sihai~Dave Zhao and Yi~Li.
\newblock Principled sure independence screening for cox models with
  ultra-high-dimensional covariates.
\newblock {\em J Multivar Anal.}, 105(1):397--411, 2012.

\bibitem{Hastie1990Generalized}
T.~Hastie and R.~Tibshirani.
\newblock {\em Generalized additive models}.
\newblock London: Chapmand and Hall, 1990.

\bibitem{Du2010PENALIZED}
P.~Du, S.~Ma, and H.~Liang.
\newblock Penalized variable selection procedure for cox models with
  semiparametric relative risk.
\newblock {\em Ann Stat.}, 38(4):2092--2117, 2010.

\bibitem{Lian2013Shrinkage}
Heng Lian, Jianbo Li, and Yuao Hu.
\newblock Shrinkage variable selection and estimation in proportional hazards
  models with additive structure and high dimensionality.
\newblock {\em Comput Stat Data Anal.}, 63(48):99--112, 2013.

\bibitem{liu2007semiparametric}
Dawei Liu, Xihong Lin, and Debashis Ghosh.
\newblock Semiparametric regression of multidimensional genetic pathway data:
  least-squares kernel machines and linear mixed models.
\newblock {\em Biometrics.}, 63(4):1079--1088, 2007.

\bibitem{liu2008estimation}
Dawei Liu, Debashis Ghosh, and Xihong Lin.
\newblock Estimation and testing for the effect of a genetic pathway on a
  disease outcome using logistic kernel machine regression via logistic mixed
  models.
\newblock {\em BMC bioinformatics.}, 9(1):1--11, 2008.

\bibitem{maity2011powerful}
Arnab Maity and Xihong Lin.
\newblock Powerful tests for detecting a gene effect in the presence of
  possible gene--gene interactions using garrote kernel machines.
\newblock {\em Biometrics.}, 67(4):1271--1284, 2011.

\bibitem{Allen2013Automatic}
Genevera~I. Allen.
\newblock Automatic feature selection via weighted kernels and regularization.
\newblock {\em J Comput Graph Stat.}, 22(2):284--299, 2013.

\bibitem{He2016Prioritizing}
Qianchuan He, Tianxi Cai, Yang Liu, Ni~Zhao, Quaker~E Harmon, Lynn~M Almli,
  Elisabeth~B Binder, Stephanie~M Engel, Kerry~J Ressler, Karen~N Conneely,
  Xihong Lin, and Michael~C. Wu.
\newblock Prioritizing individual genetic variants after kernel machine testing
  using variable selection.
\newblock {\em Genet Epidemiol.}, 40(8):722--731, 2016.

\bibitem{Rong2018More}
Yaohua Rong, Sihai~Dave Zhao, Ji~Zhu, Wei Yuan, Weihu Cheng, and Yi~Li.
\newblock More accurate semiparametric regression in pharmacogenomics.
\newblock {\em Stat Interface.}, 11(4):573--580, 2018.

\bibitem{Zheng2021A}
Xia Zheng, Yaohua Rong, Liu Ling, and Weihu Cheng.
\newblock A more accurate estimation of semiparametric logistic regression.
\newblock {\em Mathematics.}, 11(4):573--580, 2021.

\bibitem{Leng2006Model}
Chenlei Leng and Hao~Helen Zhang.
\newblock Model selection in nonparametric hazard regression.
\newblock {\em J Nonparametr Stat.}, 18(7-8):417--429, 2006.

\bibitem{Lin2006Component}
Y.~Lin and H.~H. Zhang.
\newblock Component selection and smoothing in multivariate nonparametric
  regression.
\newblock {\em Ann Stat.}, 34(5):2272--2297, 2006.

\bibitem{Li2003Kernel}
Hongzhe Li and Yihui Luan.
\newblock Kernel cox regression models for linking gene expression profiles to
  censored survival data.
\newblock {\em Pac Symp Biocomput.}, 8(12):65--76, 2003.

\bibitem{Sinnott2018Pathway}
Jennifer~A. Sinnott and Tianxi Cai.
\newblock Pathway aggregation for survival prediction via multiple kernel
  learning.
\newblock {\em Stat Med.}, 37(16):2501--2515, 2018.

\bibitem{cristianini2000introduction}
Nello Cristianini and John Shawe-Taylor.
\newblock {\em An introduction to support vector machines and other
  kernel-based learning methods}.
\newblock Cambridge: Cambridge University Press, 2000.

\bibitem{schaid2010genomic}
Daniel~J. Schaid.
\newblock Genomic similarity and kernel methods i: advancements by building on
  mathematical and statistical foundations.
\newblock {\em Hum Hered.}, 70:109--131, 2010.

\bibitem{buhmann2003radial}
Martin~D Buhmann.
\newblock {\em Radial basis functions: theory and implementations}.
\newblock Cambridge: Cambridge university press, 2003.

\bibitem{Steinwart2002On}
Ingo Steinwart.
\newblock On the influence of the kernel on the consistency of support vector
  machines.
\newblock {\em J Mach Learn Res.}, 2:67--93, 2001.

\bibitem{scholkopf2002learning}
Bernhard Sch{\"o}lkopf and Alexander~J Smola.
\newblock {\em Learning with kernels}.
\newblock Cambridge: MIT Press, 2002.

\bibitem{wu2011rare}
Michael~C Wu, Seunggeun Lee, Tianxi Cai, Yun Li, Michael Boehnke, and Xihong
  Lin.
\newblock Rare-variant association testing for sequencing data with the
  sequence kernel association test.
\newblock {\em Am J Hum Genet.}, 89(1):82--93, 2011.

\bibitem{kimeldorf1970spline}
George Kimeldorf and Grace Wahba.
\newblock Some results on tchebycheffian spline functions.
\newblock {\em J Math Anal Appl.}, 33(1):82--95, 1971.

\bibitem{friedman2007pathwise}
Jerome Friedman, Trevor Hastie, Holger H{\"o}fling, and Robert Tibshirani.
\newblock Pathwise coordinate optimization.
\newblock {\em Ann Appl Stat.}, 1(2):302--332, 2007.

\bibitem{friedman2010regularization}
Jerome Friedman, Trevor Hastie, and Rob Tibshirani.
\newblock Regularization paths for generalized linear models via coordinate
  descent.
\newblock {\em J Stat Softw.}, 33(1):1--22, 2010.

\bibitem{2009BB}
Ravi Varadhan and Paul Gilbert.
\newblock Bb: An r package for solving a large system of nonlinear equations
  and for optimizing a high-dimensional nonlinear objective function.
\newblock {\em J Stat Softw.}, 32(4):1–26, 2010.

\bibitem{Van2006Cross}
Hans~C. Van~Houwelingen, Tako Bruinsma, Augustinus~AM Hart, Laura~J.
  Van't~Veer, and Lodewyk~FA Wessels.
\newblock Cross-validated cox regression on microarray gene expression data.
\newblock {\em Stat Med.}, 25(18):3201--3216, 2006.

\bibitem{Simon2011Regularization}
Noah Simon, Jerome Friedman, Trevor Hastie, and Robert Tibshirani.
\newblock Regularization paths for cox's proportional hazards model via
  coordinate descent.
\newblock {\em J Stat Softw.}, 39(5):1--13, 2011.

\bibitem{uno2011on}
Hajime Uno, Tianxi Cai, Michael~J Pencina, Ralph~B D'Agostino, and Lee~Jen Wei.
\newblock On the c-statistics for evaluating overall adequacy of risk
  prediction procedures with censored survival data.
\newblock {\em Stat Med.}, 30(10):1105--1117, 2011.

\bibitem{heagerty2005survival}
Patrick~J Heagerty and Yingye Zheng.
\newblock Survival model predictive accuracy and {ROC} curves.
\newblock {\em Biometrics.}, 61(1):92--105, 2005.

\bibitem{broyl2010gene}
Annemiek Broyl, Dirk Hose, Henk Lokhorst, Yvonne de~Knegt, Justine Peeters,
  Anna Jauch, Uta Bertsch, Arjan Buijs, Marian Stevens-Kroef, H~Berna Beverloo,
  et~al.
\newblock Gene expression profiling for molecular classification of multiple
  myeloma in newly diagnosed patients.
\newblock {\em Blood.}, 116(14):2543--2553, 2010.

\bibitem{li2021bioinformatics}
Feng Li, Yi~Jin, Xiaolu Pei, Peiyuan Guo, Keqin Dong, Haoyuan Wang, Yujia Chen,
  Peng Guo, Ling-Bing Meng, and Zhiyu Wang.
\newblock Bioinformatics analysis and verification of gene targets for renal
  clear cell carcinoma.
\newblock {\em Comput Biol Chem.}, 92:107453, 2021.

\bibitem{williams2013oncogenic}
Sarah~V Williams, Carolyn~D Hurst, and Margaret~A Knowles.
\newblock Oncogenic fgfr3 gene fusions in bladder cancer.
\newblock {\em Hum Mol Genet.}, 22(4):795--803, 2013.

\bibitem{rushworth2014dual}
Linda~K Rushworth, Andrew~M Kidger, Laurent Delavaine, Graeme Stewart, Susanne
  Van~Schelven, Jane Davidson, Christopher~J Bryant, Edward Caddye, Philip
  East, Christopher~J Caunt, et~al.
\newblock Dual-specificity phosphatase 5 regulates nuclear erk activity and
  suppresses skin cancer by inhibiting mutant harvey-ras (hrasq61l)-driven
  serpinb2 expression.
\newblock {\em Proc Natl Acad Sci.}, 111(51):18267--18272, 2014.

\end{thebibliography}



\section*{APPENDIX}
\subsection*{$\ell^{'}_n(\bm\eta)$ and $\ell^{''}_n(\bm\eta)$}

To update the values of $\bm\beta$ and $\bm\alpha$, we  compute $\ell^{'}_n(\bm\eta)$ and $\ell^{''}_n(\bm\eta)$ in Section \ref{algo} as follows. Let $\bm\eta=(\eta_1,\eta_2,\ldots,\eta_n)^T$, where $\eta_i=\bm X_i^T\bm \beta + \bm k_i^T(\bm \delta) \bm\alpha.$ The log-partial likelihood function can be written as
\[
\ell_n(\bm\eta) = \frac{1}{n} \sum_{j=1}^n \tau_j\{\eta_j - \log[\sum_{l\in R_j} \exp(\eta_l)]\}.
\]
Let $E_i = \{R_m:i\in R_m\}$ denote the risk sets containing individual $i$, and $E_{ii^{'}} = \{R_m:i,i^{'} \in R_m\}$ be the risk sets containing both individuals $i$ and $i^{'}$. Straightforward calculations yield
\[
\frac{\partial \ell_n(\bm\eta)}{\partial \eta_i} = \frac{1}{n}[ \tau_i - \exp(\eta_i) \sum_{m\in E_i} \frac{\tau_m}{\sum_{l\in R_m}\exp(\eta_l)}],
\]
\[
\frac{\partial^{2} \ell_n(\bm\eta)}{\partial \eta_i^{2}} =  - \frac{1}{n} \exp(\eta_i) \sum_{m\in E_i} \frac{\tau_m}{\sum_{l\in R_m}\exp(\eta_l)} + \frac{1}{n} \exp(2\eta_i) \sum_{m\in E_i} \frac{\tau_m}{\{\sum_{l\in R_m}\exp(\eta_l)\}^2},
\]
\[
\frac{\partial^{2} \ell_n(\bm\eta)}{\partial \eta_i \partial \eta_{i^{'}}} = \frac{1}{n} \exp(\eta_i) \exp(\eta_{i^{'}}) \sum_{m\in E_{ii^{'}}} \frac{\tau_m}{\{\sum_{l\in R_m}\exp(\eta_l)\}^2},\ \  (i\neq i^{'}).
\]

\subsection*{Gradient with respect to $\bm \delta$}

Assuming no ties in the observed failure times, we arrange them into an ascending order. Since updating of $\bm \delta$ may involve large-scale optimization, we need to derive the gradient of objective function \eqref{maxvec} with respect to $\bm \delta$ to  accelerate the implementation of the  spectral projected gradient method.

We first recall objective function \eqref{maxvec},

\begin{equation*}
	\begin{aligned}
		\, f(\bm \alpha,\bm \beta,\bm \delta) =
		\,  & \frac{1}{n} \bm\tau^T \left[\bm X \bm\beta+ \bm K(\bm\delta)\bm\alpha - \bm A(\bm\alpha,\bm\beta,\bm\delta)  \right]        \\
		&
		- \lambda_1 \Vert\bm\beta\Vert_1 - \lambda_2\Vert\bm\delta\Vert_1 -
		\frac{1}{2}\lambda_3\bm\alpha^T \bm K(\bm\delta)\bm\alpha.
	\end{aligned}
\end{equation*}

	Denote
	$\uppercase\expandafter{\romannumeral1} = \frac{\partial \boldsymbol{\tau}^T \boldsymbol{K(\delta)}  \boldsymbol{\alpha}}{\partial \boldsymbol{\delta}},$
	$\uppercase\expandafter{\romannumeral2} = \frac{\partial \boldsymbol{\tau}^T \boldsymbol{A}(\boldsymbol{\alpha},\boldsymbol{\beta},\boldsymbol{\delta})}{\partial \boldsymbol{\delta}}{ },$
		$\uppercase\expandafter{\romannumeral3}= 
	\frac{\partial \boldsymbol{\alpha}^T \boldsymbol{K}(\boldsymbol{\delta}) \boldsymbol{\alpha}}{\partial \boldsymbol{\delta}}$ and $\uppercase\expandafter{\romannumeral1}$, $\uppercase\expandafter{\romannumeral2}$ and $\uppercase\expandafter{\romannumeral3}$ are calculated in the following \textit{Proof.} $\bm 1$, \textit{Proof.} $\bm 2$ and \textit{Proof.} $\bm 3$, respectively.
	
	\begin{proof}{$\textbf{ 1. } $ }
	\begin{flalign}
	{\qquad}& \uppercase\expandafter{\romannumeral1}=\frac{\partial \boldsymbol{\tau}^T \boldsymbol{K(\delta)}  \boldsymbol{\alpha}}{\partial \boldsymbol{\delta}} _{Q \times 1}=\left(\frac{\partial \boldsymbol{K(\delta)}}{\partial \boldsymbol{\delta}}\right)_{Q \times n n}^T\left(\frac{\partial \boldsymbol{\tau}^T \boldsymbol{K(\delta)} \boldsymbol{\alpha}}{\partial \boldsymbol{K(\delta)}}\right)_{n n \times 1}^T=\left(\frac{\partial \boldsymbol{K(\delta)}}{\partial \boldsymbol{\delta}}\right)^T\left(\boldsymbol{\tau}^T \otimes \boldsymbol{\alpha}^T\right)^T, &
		\nonumber
	\end{flalign}
where $\otimes$ represents “Kronecker product” and $\left(\frac{\partial \boldsymbol{K(\delta)}}{\partial \boldsymbol{\delta}}\right)^T$ can be expressed as follows. 
	
\begin{flalign}
	{\qquad} & \left(\frac{\partial \boldsymbol{K(\delta)}}{\partial \boldsymbol{\delta}}\right)^T=\left(\left(\frac{\partial \boldsymbol{K_1(\delta)}}{\partial \boldsymbol{\delta}}\right)^T,\left(\frac{\partial \boldsymbol{K_2(\delta)}}{\partial \boldsymbol{\delta}}\right)^T, \cdots \cdots,\left(\frac{\partial \boldsymbol{K_n(\delta)}}{\partial \boldsymbol{\delta}}\right)^T\right)_{Q \times n n} &
		\nonumber
\end{flalign}
	
\begin{flalign}
	&\ =\left(\begin{array}{ccccc}
		 \exp \left\{-\sum_{q=1}^Q \delta_q\left(z_{1 q}-z_{1 q}\right)^2\right\}\left(-\left(z_{11}-z_{11}\right)^2\right) & \ldots & \exp \left\{-\sum_{q=1}^Q \delta_q\left(z_{1 q}-z_{n q}\right)^2\right\}\left(-\left(z_{11}-z_{n 1}\right)^2\right) & \ldots \\
		\vdots & \vdots & \vdots \\
		\vdots & \vdots & \vdots \\
		\exp \left\{-\sum_{q=1}^Q \delta_q\left(z_{1 q}-z_{1 q}\right)^2\right\}\left(-\left(z_{1 Q}-z_{1 Q}\right)^2\right) & \ldots & \exp \left\{-\sum_{q=1}^Q \delta_q\left(z_{1 q}-z_{n q}\right)^2\right\}\left(-\left(z_{1 Q}-z_{n Q}\right)^2\right) & \ldots 
	\end{array}\right.  &
        \nonumber
\end{flalign}
	
\begin{flalign}
&\ {\quad}	\left.\begin{array}{ccccc}
		  \exp \left\{-\sum_{q=1}^Q \delta_q\left(z_{n q}-z_{1 q}\right)^2\right\}\left(-\left(z_{n 1}-z_{11}\right)^2\right) & \ldots & \exp \left\{-\sum_{q=1}^Q \delta_q\left(z_{n q}-z_{n q}\right)^2\right\}\left(-\left(z_{n 1}-z_{n 1}\right)^2\right) \\
		\vdots&\vdots & \vdots \\
		\vdots&\vdots & \vdots \\
		 \exp \left\{-\sum_{q=1}^Q \delta_q\left(z_{n q}-z_{1 q}\right)^2\right\}\left(-\left(z_{n Q}-z_{1 Q}\right)^2\right) & \ldots & \exp \left\{-\sum_{q=1}^Q \delta_q\left(z_{n q}-z_{n q}\right)^2\right\}\left(-\left(z_{n Q}-z_{n Q}\right)^2\right)
	\end{array}\right)_{Q \times n n}  &
\nonumber
\end{flalign}
	
\begin{flalign}
&\	=\left(\begin{array}{cccccc}
		\exp \left\{-\sum_{q=1}^Q \delta_q\left(z_{1 q}-z_{1 q}\right)^2\right\} & \ldots & \exp \left\{-\sum_{q=1}^Q \delta_q\left(z_{1 q}-z_{n q}\right)^2\right\} & \ldots & \exp \left\{-\sum_{q=1}^Q \delta_q\left(z_{n q}-z_{1 q}\right)^2\right\} & \ldots\\
		\vdots & \vdots & \vdots & \vdots& \vdots  \\
		\vdots & \vdots & \vdots & \vdots& \vdots  \\
		\exp \left\{-\sum_{q=1}^Q \delta_q\left(z_{1 q}-z_{1 q}\right)^2\right\} & \ldots & \exp \left\{-\sum_{q=1}^Q \delta_q\left(z_{1 q}-z_{n q}\right)^2\right\} & \ldots & \exp \left\{-\sum_{q=1}^Q \delta_q\left(z_{n q}-z_{1 q}\right)^2\right\} & \ldots
	\end{array}\right.  &
\nonumber
\end{flalign}

\begin{flalign}	
&\ {\quad}	{\left. {\begin{array}{*{20}{c}}
				{  {\rm{  }}\exp \left\{ { - \sum\limits_{q = 1}^Q {{\delta _q}} {{\left( {{z_{nq}} - {z_{nq}}} \right)}^2}} \right\}}\\
				\vdots \\
				\vdots \\
				{  {\rm{ }}\exp \left\{ { - \sum\limits_{q = 1}^Q {{\delta _q}} {{\left( {{z_{nq}} - {z_{nq}}} \right)}^2}} \right\}}
		\end{array}} \right)_{Q \times nn}} &
  \nonumber 
\end{flalign}

  \begin{flalign}	
&\  * {\quad}{\left( {\begin{array}{*{20}{c}}
				{ - {{\left( {{z_{11}} - {z_{11}}} \right)}^2}}& \cdots &{ - {{\left( {{z_{11}} - {z_{n1}}} \right)}^2}}& \cdots &{ - {{\left( {{z_{n1}} - {z_{11}}} \right)}^2}}& \cdots &{ - {{\left( {{z_{n1}} - {z_{n1}}} \right)}^2}}\\
				\vdots & \vdots & \vdots & \vdots & \vdots & \vdots & \vdots \\
				\vdots & \vdots & \vdots & \vdots & \vdots & \vdots & \vdots \\
				{ - {{\left( {{z_{1Q}} - {z_{1Q}}} \right)}^2}}& \cdots &{ - {{\left( {{z_{1Q}} - {z_{nQ}}} \right)}^2}}& \cdots &{ - {{\left( {{z_{nQ}} - {z_{1q}}} \right)}^2}}& \cdots &{ - {{\left( {{z_{nQ}} - {z_{nQ}}} \right)}^2}}
		\end{array}} \right)_{Q \times nn}}
   &
  \nonumber 
\end{flalign}
	
	\begin{flalign}
	&\	=\left(\boldsymbol{E}_{Q \times 1}(\operatorname{Vec}(\boldsymbol{K}(\boldsymbol{\delta})))^T * \boldsymbol{B}\right)_{Q \times n n} &
	\nonumber
	\end{flalign}
where $\boldsymbol{E}$ is a $Q \times 1$ vector whose elements are $1$, $\operatorname{Vec} (\cdot)$ is a matrix vec operator,  $*$ represents hadamard product and  $\boldsymbol{B}=(-u[,, 1],-u[,, 2], \cdots,-u[,, n])$ is a ${Q \times n n}$ matrix where $ - u[,,1] $, $ - u[,,2] $, $\cdots$, $ - u[,,n] $ are expressed as

    \[ - u[,,1] = {\left( {\begin{array}{*{20}{c}}
    			{ - {{\left( {{z_{11}} - {z_{11}}} \right)}^2}}& \cdots &{ - {{\left( {{z_{11}} - {z_{n1}}} \right)}^2}}\\
    			\vdots & \vdots & \vdots \\
    			\vdots & \vdots & \vdots \\
    			{ - {{\left( {{z_{1Q}} - {z_{1Q}}} \right)}^2}}& \cdots &{ - {{\left( {{z_{1Q}} - {z_{nQ}}} \right)}^2}}
    	\end{array}} \right)_{Q \times n}},{\rm{      }}\]
    	
        \[ - u[,,2] = {\left( {\begin{array}{*{20}{c}}
    			{ - {{\left( {{z_{21}} - {z_{11}}} \right)}^2}}& \cdots &{ - {{\left( {{z_{21}} - {z_{n1}}} \right)}^2}}\\
    			\vdots & \vdots & \vdots \\
    			\vdots & \vdots & \vdots \\
    			{ - {{\left( {{z_{2Q}} - {z_{1Q}}} \right)}^2}}& \cdots &{ - {{\left( {{z_{2Q}} - {z_{nQ}}} \right)}^2}}
    	\end{array}} \right)_{Q \times n}},{\rm{      }}\]
    	
    	\[ \vdots\]
    	
    \[ - u[,,n] = {\left( {\begin{array}{*{20}{c}}
    			{ - {{\left( {{z_{n1}} - {z_{11}}} \right)}^2}}& \cdots &{ - {{\left( {{z_{n1}} - {z_{n1}}} \right)}^2}}\\
    			\vdots & \vdots & \vdots \\
    			\vdots & \vdots & \vdots \\
    			{ - {{\left( {{z_{nQ}} - {z_{1Q}}} \right)}^2}}& \cdots &{ - {{\left( {{z_{nQ}} - {z_{nQ}}} \right)}^2}}
    	\end{array}} \right)_{Q \times n}}.\]

In summary, $ \uppercase\expandafter{\romannumeral1} = \frac{\partial \boldsymbol{\tau}^T \boldsymbol{K}(\boldsymbol{\delta}) \boldsymbol{\alpha}}{\partial \boldsymbol{\delta}}{ } $ can be written as

    \begin{flalign}
    	\uppercase\expandafter{\romannumeral1} = \frac{\partial \boldsymbol{\tau}^T \boldsymbol{K}(\boldsymbol{\delta}) \boldsymbol{\alpha}}{\partial \boldsymbol{\delta}}{ }_{Q \times 1}=\left(\boldsymbol{E}_{Q \times 1}(\operatorname{Vec}(\boldsymbol{K}(\boldsymbol{\delta})))^T * \boldsymbol{B}\right)_{Q \times n n}(\boldsymbol{\tau} \otimes \boldsymbol{\alpha})_{n n \times 1}. 
    	\nonumber
    \end{flalign}
    
\end{proof}


    \thispagestyle{empty}
\begin{proof}{$\textbf{ 2. } $ }	
	\begin{flalign}
	{\quad} \uppercase\expandafter{\romannumeral2} = \frac{\partial \boldsymbol{\tau}^T \boldsymbol{A}(\boldsymbol{\alpha},\boldsymbol{\beta},\boldsymbol{\delta})}{\partial \boldsymbol{\delta}}_{Q \times 1}=\left(\frac{\partial \boldsymbol{K}(\boldsymbol{\delta})}{\partial \boldsymbol{\delta}}\right)_{Q \times n n}^T\left(\frac{\partial \boldsymbol{\tau}^T \boldsymbol{A}(\boldsymbol{\alpha},\boldsymbol{\beta},\boldsymbol{\delta})}{\partial \boldsymbol{K}(\boldsymbol{\delta})}\right)_{n n \times 1}^T 
\nonumber
\end{flalign}  

\begin{flalign}
  =\left(\boldsymbol{E}_{Q \times 1}(\operatorname{Vec}(\boldsymbol{K}(\boldsymbol{\delta})))^T * \boldsymbol{B}\right)_{Q \times n n}\left(\frac{\partial \boldsymbol{\tau}^T \boldsymbol{A}(\boldsymbol{\alpha},\boldsymbol{\beta},\boldsymbol{\delta})}{\partial \boldsymbol{K}(\boldsymbol{\delta})}\right)_{n n \times 1}^T. 
			\nonumber
		\end{flalign}
 $\frac{\partial \boldsymbol{\tau}^T \boldsymbol{A}(\boldsymbol{\alpha},\boldsymbol{\beta},\boldsymbol{\delta})}{\partial \boldsymbol{K}(\boldsymbol{\delta})}$ can be expressed as follows,

		\begin{flalign}
		{\qquad}	& \frac{\partial \boldsymbol{\tau}^T \boldsymbol{A}(\boldsymbol{\alpha},\boldsymbol{\beta},\boldsymbol{\delta})}{\partial \boldsymbol{K}(\boldsymbol{\delta})}=\left(\left(\frac{\partial \boldsymbol{\tau}^T \boldsymbol{A}(\boldsymbol{\alpha},\boldsymbol{\beta},\boldsymbol{\delta})}{\partial \boldsymbol{k_1}(\boldsymbol{\delta})}\right),\left(\frac{\partial \boldsymbol{\tau}^T \boldsymbol{A}(\boldsymbol{\alpha},\boldsymbol{\beta},\boldsymbol{\delta})}{\partial \boldsymbol{k}_2(\boldsymbol{\delta})}\right), \cdots \cdots,\left(\frac{\partial \boldsymbol{\tau}^T \boldsymbol{A}(\boldsymbol{\alpha},\boldsymbol{\beta},\boldsymbol{\delta})}{\partial \boldsymbol{k_n}(\boldsymbol{\delta})}\right)\right)_{1 \times n n}&
			\nonumber
	\end{flalign}

		\begin{flalign}
& {\quad}	=\left(\tau_1 \frac{\left[\exp \left(\boldsymbol{X}_1^T \boldsymbol{\beta}+\boldsymbol{k}_1(\boldsymbol{\delta})\right)\right] \boldsymbol{\alpha}^T}{\sum_{l \in R_1} \exp \left(\boldsymbol{X}_l^T \boldsymbol{\beta}+\boldsymbol{k}_l(\boldsymbol{\delta}) \boldsymbol{\alpha}\right)}, \tau_1 \frac{\left[\exp \left(\boldsymbol{X}_2^T \boldsymbol{\beta}+\boldsymbol{k}_2(\boldsymbol{\delta})\right)\right] \boldsymbol{\alpha}^T}{\sum_{l \in R_1} \exp \left(\boldsymbol{X}_l^T \boldsymbol{\beta}+\boldsymbol{k}_l(\boldsymbol{\delta}) \boldsymbol{\alpha}\right)}+\tau_2 \frac{\left[\exp \left(\boldsymbol{X}_2^T \boldsymbol{\beta}+\boldsymbol{k}_2(\boldsymbol{\delta})\right)\right] \boldsymbol{\alpha}^T}{\sum_{l \in R_2} \exp \left(\boldsymbol{X}_l^T \boldsymbol{\beta}+\boldsymbol{k}_l(\boldsymbol{\delta}) \boldsymbol{\alpha}\right)}, \cdots\right)_{1 \times n n} &
			\nonumber
\end{flalign}


  \begin{flalign}
	{\quad} 	& =\bm F  * {\left( {{\boldsymbol{\alpha} ^T},{\boldsymbol{\alpha} ^T}, \cdots ,{\boldsymbol{\alpha} ^T}} \right)_{1\times nn}}, &
  \nonumber
  \end{flalign}  
where $\bm F= (g_1,\cdots,g_1,g_2,\cdots,g_2,\cdots,g_n,\cdots,g_n)$ is a $1 \times nn$ vector and  $g_i$ $(i=1,2,\cdots n)$ are repeated $n$ times, respectively. 
$$
g_i= \tau_1 \frac{\left[\exp \left(\boldsymbol{X}_i^T \boldsymbol{\beta}+\boldsymbol{k}_i(\boldsymbol{\delta})\right)\right]}{\sum_{l \in R_1} \exp \left(\boldsymbol{X}_l^T \boldsymbol{\beta}+\boldsymbol{k}_l(\boldsymbol{\delta}) \boldsymbol{\alpha}\right)}+ \cdots + \tau_i \frac{\left[\exp \left(\boldsymbol{X}_i^T \boldsymbol{\beta}+\boldsymbol{k}_i(\boldsymbol{\delta})\right)\right]}{\sum_{l \in R_i} \exp \left(\boldsymbol{X}_l^T \boldsymbol{\beta}+\boldsymbol{k}_l(\boldsymbol{\delta}) \boldsymbol{\alpha}\right)},\qquad i=1,2,\cdots,n.
$$ 

Denote $\bm G = (g_1,\cdots,g_n)$, which can be written as, 	
 
 \begin{flalign}
 {\quad}\bm G=	&\  \left(\tau_1 \frac{\left[\exp \left(\boldsymbol{X}_1^T \boldsymbol{\beta}+\boldsymbol{k}_1(\boldsymbol{\delta})\right)\right]}{\sum_{l \in R_1} \exp \left(\boldsymbol{X}_l^T \boldsymbol{\beta}+\boldsymbol{k}_l(\boldsymbol{\delta}) \boldsymbol{\alpha}\right)}, \tau_1 \frac{\left[\exp \left(\boldsymbol{X}_2^T \boldsymbol{\beta}+\boldsymbol{k}_2(\boldsymbol{\delta})\right)\right]}{\sum_{l \in R_1} \exp \left(\boldsymbol{X}_l^T \boldsymbol{\beta}+\boldsymbol{k}_l(\boldsymbol{\delta}) \boldsymbol{\alpha}\right)}+\tau_2 \frac{\left[\exp \left(\boldsymbol{X}_2^T \boldsymbol{\beta}+\boldsymbol{k}_2(\boldsymbol{\delta})\right)\right]}{\sum_{l \in R_2} \exp \left(\boldsymbol{X}_l^T \boldsymbol{\beta}+\boldsymbol{k}_l(\boldsymbol{\delta}) \boldsymbol{\alpha}\right)}, \cdots\right)_{1 \times n} &
 	\nonumber
 \end{flalign}

\begin{flalign}
	& {\quad}=\left(\frac{\tau_1}{\sum_{l \in R_1} \exp \left(\boldsymbol{X}_l^T \boldsymbol{\beta}+\boldsymbol{k}_l(\boldsymbol{\delta}) \boldsymbol{\alpha}\right)}, \frac{\tau_1}{\sum_{l \in R_1} \exp \left(\boldsymbol{X}_l^T \boldsymbol{\beta}+\boldsymbol{k}_l(\boldsymbol{\delta}) \boldsymbol{\alpha}\right)}+\frac{\tau_2}{\sum_{l \in R_2} \exp \left(\boldsymbol{X}_l^T \boldsymbol{\beta}+\boldsymbol{k}_l(\boldsymbol{\delta}) \boldsymbol{\alpha}\right)}, \cdots  \right)_{1 \times n} &
	\nonumber
\end{flalign}
\begin{flalign}
	  {\qquad}	&\ *\left(\exp \left(\boldsymbol{X}_1^T \boldsymbol{\beta}+\boldsymbol{k}_1(\boldsymbol{\delta})\right), \exp \left(\boldsymbol{X}_2^T \boldsymbol{\beta}+\right.\right. \left.\left.\boldsymbol{k}_2(\boldsymbol{\delta})\right), \cdots  \right)_{1 \times n} &
\nonumber
\end{flalign}

\begin{flalign}
	&\ {\quad} =\left(\frac{\tau_1}{\sum_{l \in R_1} \exp \left(\boldsymbol{X}_l^T \boldsymbol{\beta}+\boldsymbol{k}_l(\boldsymbol{\delta}) \boldsymbol{\alpha}\right)}, \frac{\tau_2}{\sum_{l \in R_2} \exp \left(\boldsymbol{X}_l^T \boldsymbol{\beta}+\boldsymbol{k}_l(\boldsymbol{\delta}) \boldsymbol{\alpha}\right)}, \cdots \right)_{1 \times n} {\rm{  }} \left(\begin{array}{llll}
		1 & 1 & \cdots & 1      \\
		0 & 1 & \cdots & 1      \\
		\vdots  & \vdots   & \ddots       & \vdots \\
		 0 & 0  & \cdots       & 1
	\end{array}\right)_{n \times n} &
	\nonumber
\end{flalign}

\begin{flalign}
 {\qquad}	&\ *\left( \exp \left(\boldsymbol{X}_1^T \boldsymbol{\beta}+\boldsymbol{k}_1(\boldsymbol{\delta})\right), \exp \left(\boldsymbol{X}_2^T \boldsymbol{\beta}+\boldsymbol{k}_2(\boldsymbol{\delta})\right), \cdots \right)_{1 \times n}. &
	\nonumber
\end{flalign}

The $\bm G$ is derived for facilitating calculation of $\bm F$ and  $\uppercase\expandafter{\romannumeral2}$  can be obtained,

\begin{equation}
	\uppercase\expandafter{\romannumeral2} = \frac{\partial \boldsymbol{\tau}^T \boldsymbol{A}(\boldsymbol{\alpha},\boldsymbol{\beta},\boldsymbol{\delta})}{\partial \boldsymbol{\delta}}_{Q \times 1}=\left(\boldsymbol{E}_{Q \times 1}(\operatorname{Vec}(\boldsymbol{K}(\boldsymbol{\delta})))^T * \boldsymbol{B}\right)_{Q \times n n}\left( \bm F  * {\left( {{\boldsymbol{\alpha} ^T},{\boldsymbol{\alpha} ^T}, \cdots ,{\boldsymbol{\alpha} ^T}} \right)} \right)_{n n \times 1}.
	\nonumber
\end{equation}
\end{proof}	


\begin{proof}{$\textbf{ 3. } $ }
Similar to $\uppercase\expandafter{\romannumeral1} $, $\uppercase\expandafter{\romannumeral3}$
can be written as,
	
    $$
	\frac{\partial \boldsymbol{\alpha}^T \boldsymbol{K}(\boldsymbol{\delta}) \boldsymbol{\alpha}}{\partial \boldsymbol{\delta}}{ }_{Q \times 1}=\left(\boldsymbol{E}_{Q \times 1}(\operatorname{Vec}(\boldsymbol{K}(\boldsymbol{\delta})))^T * \boldsymbol{B}\right)_{Q \times n n}(\boldsymbol{\alpha} \otimes \boldsymbol{\alpha})_{n n \times 1}.
	$$
\end{proof}


Combining \textit{Proof.} $\bm 1$, \textit{Proof.} $\bm 2$ and \textit{Proof.} $\bm 3$, we obtain the following results,
	
	\begin{equation}
		\begin{aligned}
			& \frac{\partial f(\boldsymbol{\alpha}, \boldsymbol{\beta}, \boldsymbol{\delta})}{\partial \boldsymbol{\delta}}= \frac{1}{n}\left[\left(\boldsymbol{E}_{Q \times 1}(\operatorname{Vec}(\boldsymbol{K}(\boldsymbol{\delta})))^T * \boldsymbol{B}\right)_{Q \times n n}(\boldsymbol{\tau} \otimes \boldsymbol{\alpha})_{n n \times 1} \right.\\
  & {\qquad}	{\qquad}	{\qquad} \left. -\left(\boldsymbol{E}_{Q \times 1}(\operatorname{Vec}(\boldsymbol{K}(\boldsymbol{\delta})))^T * \boldsymbol{B}\right)_{Q \times n n}\left( \bm F  * {\left( {{\boldsymbol{\alpha} ^T},{\boldsymbol{\alpha} ^T}, \cdots ,{\boldsymbol{\alpha} ^T}} \right)} \right)_{n n \times 1} \right] \\
			&	{\qquad}	{\qquad}	{\qquad} -\lambda_2 \boldsymbol{E}_{Q \times 1}-\frac{1}{2} \lambda_3\left(\boldsymbol{E}_{Q \times 1}(\operatorname{Vec}(\boldsymbol{K}(\boldsymbol{\delta})))^T * \boldsymbol{B}\right)_{Q \times n n}(\boldsymbol{\alpha} \otimes \boldsymbol{\alpha})_{n n \times 1},
		\end{aligned}
	\nonumber
	\end{equation}
where $\bm E$, $\bm B$ and $\bm F$ have been previously defined.

\clearpage

\begin{center}
\begin{table*}[t]
	\caption{Comparison of prediction performances of the RegGKM and the LASSO-COX methods under multivariate settings}
	\centering
		\begin{threeparttable}
			\begin{tabular}{cccccccc}
		\toprule
		\toprule
		CR & Setting  & n &P  & Q   &Prediction Measure & RegGKM & LASSO-COX \\ \midrule
		\multirow{6}{*}{0\%} & \multirow{3}{*}{1}& \multirow{3}{*}{100} & \multirow{3}{*}{1} & \multirow{3}{*}{5} &  CVPL        &\textbf{-2.9387} (0.0906)   & -3.3393 (0.3033)           \\
	&	&      &      &       &  C-statistics      & \textbf{0.8601} (0.0238)    & 0.8106 (0.0248)          \\
	&	&      &      &       &  AUC        &  \textbf{0.9502} (0.0349)    & 0.9291 (0.0569)          \\
		& \multirow{3}{*}{2}& \multirow{3}{*}{100} & \multirow{3}{*}{2} & \multirow{3}{*}{15} & CVPL   & \textbf{-2.9776} (0.0933)   & -3.2197 (0.0998)             \\
	&	&     &       &       & C-statistics         &  \textbf{0.8503} (0.0228)    & 0.8060 (0.0276)           \\
	&	&     &       &       &  AUC        & \textbf{0.9508} (0.0359)    & 0.9208 (0.0758)          \\ \midrule
	\multirow{6}{*}{10\%} & \multirow{3}{*}{1} & \multirow{3}{*}{100} & \multirow{3}{*}{1} & \multirow{3}{*}{5} & CVPL         & \textbf{-2.7987} (0.1225)   & -3.1099 (0.3284)           \\
	&	&     &       &        & C-statistics         &  \textbf{0.8508} (0.0243)    & 0.8081 (0.0235)            \\
	&	&     &       &        & AUC         & \textbf{0.9161} (0.0675)    & 0.8771 (0.0917)           \\
	& \multirow{3}{*}{2} & \multirow{3}{*}{100} & \multirow{3}{*}{2} & \multirow{3}{*}{15} & CVPL           & \textbf{-2.8069} (0.1279)   & -3.0079 (0.1285)           \\
	&	&     &       &        & C-statistics        & \textbf{0.8509} (0.0215)    & 0.8040 (0.0248)           \\
	&	&     &       &        &  AUC      & \textbf{0.9187} (0.0712)    & 0.8744 (0.0897)           \\ \midrule
		\multirow{6}{*}{20\%} & \multirow{3}{*}{1} & \multirow{3}{*}{100} & \multirow{3}{*}{1} & \multirow{3}{*}{5} & CVPL         & \textbf{-2.6329} (0.1593)   & -2.9271 (0.3126)           \\
	&	&     &       &        & C-statistics         & \textbf{0.8503} (0.0258)    & 0.8025 (0.0269)            \\
	&	&     &       &        & AUC         &  \textbf{0.8802} (0.1019)    & 0.8454 (0.1195)           \\
		& \multirow{3}{*}{2} & \multirow{3}{*}{100} & \multirow{3}{*}{2} & \multirow{3}{*}{15} & CVPL           & \textbf{-2.6438} (0.1286)    & -2.8353 (0.1397)           \\
	&	&     &       &        & C-statistics        & \textbf{0.8409} (0.0212)    & 0.7957 (0.0282)           \\
	&	&     &       &        &  AUC      & \textbf{0.8809} (0.1032)    & 0.8261 (0.1232)           \\ 
     \bottomrule
	\end{tabular}
\begin{tablenotes}
	\footnotesize
	\item[*] The average CVPLs, C-statistics and AUCs over 100 replications, with standard deviations in parentheses, are given in the last two columns when the censor rates equal to 0\%, 10\% and 20\%, respectively.
\end{tablenotes}
\end{threeparttable}
\label{mulvar}
\end{table*}
\end{center}

\begin{center}
\begin{table*}[t]
	\caption{Comparison of prediction performances of the RegGKM and the LASSO-COX methods under high-dimensional and ultrahigh-dimensional settings}
	\centering
		\begin{threeparttable}
	\begin{tabular}{cccccccc}
		\toprule
		\toprule
		CR                 & Setting           & n          &P       & Q                 &Prediction Measure & RegGKM & LASSO-COX \\ \midrule
		\multirow{15}{*}{0\%} 
	& \multirow{3}{*}{3}& \multirow{3}{*}{100} & \multirow{3}{*}{200} & \multirow{3}{*}{15} & CVPL   & \textbf{-3.1062} (0.0896)   & -3.5176 (0.0670)             \\
	&	&     &       &       & C-statistics         & \textbf{0.8402} (0.0254)    & 0.7341 (0.0332)           \\
	&	&     &       &       &  AUC        & \textbf{0.9687} (0.0263)    & 0.9053 (0.0771)          \\
    & \multirow{3}{*}{4} & \multirow{3}{*}{100} & \multirow{3}{*}{15} & \multirow{3}{*}{200} & CVPL          &  \textbf{-3.3618} (0.0843)        & -3.5396  (0.0552)        \\
	&	&    &       &       & C-statistics         &  \textbf{0.7498} (0.0391)        & 0.7331 (0.0293)          \\
	&	&     &       &       &  AUC        &   \textbf{0.9456} (0.0426)       &  0.9012 (0.0792)        \\
		& \multirow{3}{*}{5} & \multirow{3}{*}{100} & \multirow{3}{*}{200} & \multirow{3}{*}{200} & CVPL        &  \textbf{-3.3976} (0.0908)         & -3.5419 (0.0748)         \\
	&	&     &       &        &  C-statistics       &  \textbf{0.7583} (0.0429)         & 0.7179  (0.03341)        \\
	&	&     &       &        &  AUC      &  \textbf{0.9402} (0.0716)        &  0.8997 (0.0852)        \\ 
	& \multirow{3}{*}{6} & \multirow{3}{*}{100} & \multirow{3}{*}{1} & \multirow{3}{*}{1000} & CVPL        &  \textbf{-3.3412} (0.0872)         & -3.5593 (0.0733)         \\
	&	&     &       &        &  C-statistics       &  \textbf{0.7417} (0.0485)         & 0.7165  (0.0314)        \\
	&	&     &       &        &  AUC      &  \textbf{0.9264} (0.0575)        &  0.9088 (0.0723)        \\ 
	& \multirow{3}{*}{7} & \multirow{3}{*}{100} & \multirow{3}{*}{1000} & \multirow{3}{*}{1000} & CVPL        &  \textbf{-3.4549} (0.1035)         & -3.5813 (0.0917)         \\
	&	&     &       &        &  C-statistics       &  \textbf{0.7429} (0.0461)         & 0.7049  (0.0344)        \\
	&	&     &       &        &  AUC      &  \textbf{0.9401} (0.0762)        &  0.9014 (0.0782)        \\ \midrule
		\multirow{15}{*}{10\%} 
	& \multirow{3}{*}{3}& \multirow{3}{*}{100} & \multirow{3}{*}{200} & \multirow{3}{*}{15} & CVPL   & \textbf{-2.8910} (0.1458)   & -3.2383 (0.1380)             \\
	&	&     &       &       & C-statistics         & \textbf{0.8326} (0.0271)    & 0.7381 (0.0322)           \\
	&	&     &       &       &  AUC        & \textbf{0.9209} (0.0663)    & 0.8375 (0.0904)          \\
	& \multirow{3}{*}{4} & \multirow{3}{*}{100} & \multirow{3}{*}{15} & \multirow{3}{*}{200} &CVPL         & \textbf{-3.1813} (0.1221)        &  -3.2909 (0.1196)         \\
	&	&     &     &      &  C-statistics       & \textbf{0.7342} (0.0542)         &  0.7216 (0.0294)         \\
	&	&     &     &      & AUC        &  \textbf{0.8709} (0.1096)        &  0.8351 (0.1065)         \\
	& \multirow{3}{*}{5} & \multirow{3}{*}{100} & \multirow{3}{*}{200} & \multirow{3}{*}{200} & CVPL        &  \textbf{-3.1796} (0.1419)        & -3.2783 (0.1308)         \\
	&  &    &     &      &  C-statistics       & \textbf{0.7409} (0.0435)        &   0.7129 (0.0341)        \\
	&  &      &     &   &  AUC       & \textbf{0.8873} (0.0873)        &  0.8509 (0.0832)        \\ 
	& \multirow{3}{*}{6} & \multirow{3}{*}{100} & \multirow{3}{*}{1} & \multirow{3}{*}{1000} & CVPL        &  \textbf{-3.1596} (0.1216)        & -3.3069 (0.1321)         \\
	&  &    &     &      &  C-statistics       & \textbf{0.7239} (0.0538)        &   0.7102 (0.0352)        \\
	&  &      &     &   &  AUC       &  \textbf{0.8599} (0.0913)        &  0.8263 (0.0957)   \\     
	& \multirow{3}{*}{7} & \multirow{3}{*}{100} & \multirow{3}{*}{1000} & \multirow{3}{*}{1000} & CVPL        &  \textbf{-3.2358} (0.1539)        & -3.3221 (0.1347)         \\
	&  &    &     &      &  C-statistics       & \textbf{0.7269} (0.0573)        &   0.7029 (0.0371)        \\
	&  &      &     &   &  AUC       & \textbf{0.8678} (0.0993)        &   0.8429 (0.1095)  \\   \midrule
		\multirow{15}{*}{20\%} 
	& \multirow{3}{*}{3}& \multirow{3}{*}{100} & \multirow{3}{*}{200} & \multirow{3}{*}{15} & CVPL   & \textbf{-2.9429} (0.1372)   & -3.3058 (0.1236)             \\
	&	&     &       &       & C-statistics         & \textbf{0.8401} (0.0262)    & 0.7358 (0.0336)           \\
	&	&     &       &       &  AUC        & \textbf{0.9218} (0.0669)    & 0.8587 (0.0818)          \\
	& \multirow{3}{*}{4} & \multirow{3}{*}{100} & \multirow{3}{*}{15} & \multirow{3}{*}{200} &CVPL         &  \textbf{-2.9726}  (0.1559)       & -3.0524 (0.1496)         \\
	&	&     &      &        & C-statistics        &  \textbf{0.7189} (0.0633)       & 0.7169 (0.0360)         \\
	&	&     &      &         &  AUC       &  \textbf{0.8459} (0.0958)        &   0.7792 (0.1041)       \\
		& \multirow{3}{*}{5} & \multirow{3}{*}{100} & \multirow{3}{*}{200} & \multirow{3}{*}{200} &  CVPL       &   \textbf{-2.9597} (0.1753)        &  -3.0132 (0.1604)       \\
	&	&     &       &       &  C-statistics       &   \textbf{0.7158} (0.0667)        &  0.7133 (0.0394)        \\
	&	&     &       &       &  AUC       &  \textbf{0.8296} (0.1041)       &   0.7964 (0.1085)        \\ 
		& \multirow{3}{*}{6} & \multirow{3}{*}{100} & \multirow{3}{*}{1} & \multirow{3}{*}{1000} &  CVPL       &   \textbf{-2.9719} (0.1758)        &  -3.0178 (0.1587)       \\
	&	&     &       &       &  C-statistics       &   \textbf{0.7169} (0.0743)        &  0.7112 (0.0422)        \\
	&	&     &       &       &  AUC       &  \textbf{0.8141 } (0.1366)       &   0.7981 (0.1009)        \\ 
	& \multirow{3}{*}{7} & \multirow{3}{*}{100} & \multirow{3}{*}{1000} & \multirow{3}{*}{1000} &  CVPL       &   \textbf{-2.9558} (0.1681)        &  -3.0445 (0.1603)       \\
	&	&     &       &       &  C-statistics       &   \textbf{0.7189} (0.0492)        &  0.6943 (0.0408)        \\
	&	&     &       &       &  AUC       &  \textbf{0.8301} (0.0994)       &   0.7738 (0.1359)        \\
     \bottomrule
	\end{tabular}
\begin{tablenotes}
	\footnotesize
	\item[*] The average CVPLs, C-statistics and AUCs over 100 replications, with standard deviations in parentheses, are given in the last two columns when the censor rates equal to 0\%, 10\% and 20\%, respectively.
\end{tablenotes}
\end{threeparttable}
\label{higdim}
\end{table*}
\end{center}

\begin{center}
\begin{table}[t]
	\caption{Comparison of prediction performances of the RegGKM and the COSSO-COX methods}
	\centering
	\begin{threeparttable}
	\begin{tabular}{ccccccc}
		\toprule
		\toprule
		CR                 & Setting           & n                 & Q                 & Prediction Measure & RegGKM & COSSO-COX \\ \midrule
		\multirow{6}{*}{0\%} & \multirow{3}{*}{1} & \multirow{3}{*}{100} & \multirow{3}{*}{5} &  CVPL        & \textbf{-2.9442} (0.0851)  & -3.2851 (0.2782)          \\
		&      &      &       &  C-statistics      & \textbf{0.8530} (0.0236)   & 0.7965 (0.0325)          \\
		&      &      &       &  AUC        & \textbf{0.9466} (0.0344)   & 0.9338 (0.0651)        \\
		& \multirow{3}{*}{2} & \multirow{3}{*}{100} & \multirow{3}{*}{15} & CVPL   & \textbf{-2.9759} (0.0885)  & -3.7764 (2.2882)             \\
		&     &       &       & C-statistics         & \textbf{0.8492} (0.0202)   & 0.7821 (0.0431)           \\
		&     &       &       &  AUC        & \textbf{0.9450} (0.0381)   & 0.9022 (0.1321)            \\ \midrule

		\multirow{6}{*}{10\%} & \multirow{3}{*}{1} & \multirow{3}{*}{100} & \multirow{3}{*}{5} &  CVPL        & \textbf{-2.8098} (0.1226)  & -3.1522 (0.3518)          \\
		&      &      &       &  C-statistics      & \textbf{0.8437} (0.0245)   & 0.7845 (0.0440)          \\
		&      &      &       &  AUC        & \textbf{0.9076} (0.0671)   & 0.8431 (0.1127)        \\
		& \multirow{3}{*}{2} & \multirow{3}{*}{100} & \multirow{3}{*}{15} & CVPL   & \textbf{-2.8078} (0.1264)  & -3.9195 (2.6411)             \\
		&     &       &       & C-statistics         & \textbf{0.8449} (0.0204)   & 0.7733 (0.0542)           \\
		&     &       &       &  AUC        & \textbf{0.9122} (0.0685)   & 0.8348 (0.1581)            \\ \midrule

		\multirow{6}{*}{20\%} & \multirow{3}{*}{1} & \multirow{3}{*}{100} & \multirow{3}{*}{5} & CVPL            & \textbf{-2.6215} (0.1591)  & -3.2968 (2.6661)          \\
		&     &       &        & C-statistics        & \textbf{0.8393} (0.0257)   & 0.7752 (0.0483)         \\
		&     &       &        & AUC         & \textbf{0.8823} (0.0981)   & 0.8072 (0.1340)            \\
		& \multirow{3}{*}{2} & \multirow{3}{*}{100} & \multirow{3}{*}{15} & CVPL     & \textbf{-2.6793} (0.1524)  & -4.1370 (2.6629)         \\
		&     &       &        & C-statistics        & \textbf{0.8345} (0.0256)   & 0.7482 (0.0583)        \\
		&     &       &        &  AUC      & \textbf{0.8789} (0.0883)   & 0.7619 (0.1375)           \\
		\bottomrule
	\end{tabular}
		\begin{tablenotes}
			\footnotesize
			\item[*] The last four columns give the average CVPLs, C-statistics and AUCs over 100 replications with standard deviations in parentheses, when the censor rates equal to 0\%, 10\%, 20\%, respectively.
		\end{tablenotes}
	\end{threeparttable}
	\label{prediction4}
\end{table}
\end{center}

\begin{center}
\begin{table}[t]
	\caption[Average prediction performances]{Average prediction performances for three methods over 500 replications, with standard deviations in parentheses}
	\centering
	\begin{tabular}{cccc}
		\toprule
		\toprule
		Methods  & CVPL (SD) & C-statistics (SD) & AUC (SD) \\
		\midrule
		$\text{RegGKM}$  & \textbf{-2.0508} (0.1894)  & \textbf{0.6816} (0.0574) & \textbf{0.7446} (0.1048) \\
		$\text{COSSO-COX}$ & -2.1353 (0.2519) & 0.6396 (0.0635) & 0.6987 (0.1135) \\
		$\text{LASSO-COX}$  & -2.0850 (0.2105) &  0.6714 (0.0659) &  0.7245 (0.1196)  \\
		\bottomrule
	\end{tabular}
	\label{realtable}
\end{table}
\end{center}

\end{document}